\documentclass[3p]{elsarticle} 
\usepackage{amsmath}
\usepackage{lineno,hyperref}
\usepackage{subcaption}
\usepackage[english]{babel}
\usepackage[utf8]{inputenc}
\usepackage{algorithm}
\usepackage[noend]{algpseudocode}
\usepackage[dvipsnames]{xcolor}
\usepackage{array,booktabs,caption,ragged2e}
\usepackage{multirow}
\usepackage{pdfpages}
\usepackage{graphicx}

\journal{Journal}

\newcommand{\mc}[1]{\mathcal{#1}}
\newcommand{\LineComment}[1]{\Statex $\triangleright$ {#1}}

\usepackage{amssymb}

\begin{document}
	
	\begin{frontmatter}
		
		\title{A Multi-cascaded Model with Data Augmentation for Enhanced Paraphrase Detection in Short Texts}
		
		
		\author{Muhammad Haroon Shakeel}
		\ead{15030040@lums.edu.pk}
		\author{Asim Karim}
		\ead{akarim@lums.edu.pk}
		\author{Imdadullah Khan}
		\ead{imdad.khan@lums.edu.pk}

		
		\address{Department of Computer Science, Syed Babar Ali School of Science and Engineering, Lahore University of Management Sciences, Lahore, Pakistan}
		
		
		\cortext[mycorrespondingauthor]{Muhammad Haroon Shakeel}
		
		\begin{abstract}
			Paraphrase detection is an important task in text analytics with numerous applications such as plagiarism detection, duplicate question identification, and enhanced customer support helpdesks. Deep models have been proposed for representing and classifying paraphrases. These models, however, require large quantities of human-labeled data, which is expensive to obtain. In this work, we present a data augmentation strategy and a multi-cascaded model for improved paraphrase detection in short texts. Our data augmentation strategy considers the notions of paraphrases and non-paraphrases as binary relations over the set of texts. Subsequently, it uses graph theoretic concepts to efficiently generate additional paraphrase and non-paraphrase pairs in a sound manner. Our multi-cascaded model employs three supervised feature learners (cascades) based on CNN and LSTM networks with and without soft-attention. The learned features, together with hand-crafted linguistic features, are then forwarded to a discriminator network for final classification. Our model is both wide and deep and provides greater robustness across clean and noisy short texts. We evaluate our approach on three benchmark datasets and show that it produces a comparable or state-of-the-art performance on all three.
			
		\end{abstract}
		
		\begin{keyword}
			\texttt{paraphrase detection \sep deep learning \sep data augmentation \sep sentence similarity}
		\end{keyword}
		
	\end{frontmatter}
	
	

	\section{Introduction} \label{sec:Introduction}
	In recent years, short text in the form of posts on microblogs, question answer forums, news headlines, and tweets is being generated in abundance~\cite{cagnina2014efficient}. Performing NLP tasks is relatively easier in longer documents (e.g. news articles) than in short texts (e.g. headlines) because, in longer documents, greater context is available for semantic understanding~\cite{rashid2019fuzzy}. Moreover, in many cases, short texts (e.g. tweets) tend to use informal language (spelling variations, improper grammar, slang) compared to longer documents (e.g. blogs). Thus, the techniques tailored for formal and clean text do not perform well on informal one~\cite{dey2016paraphrase}, which call for a need to develop an approach that can work in both settings (i.e., clean and noisy informal text)~\cite{agarwal2018deep}.
	
	 A paraphrase of a document is another document that can be different in syntax, but that expresses the same meaning in the same language. Automatically detecting paraphrases among a set of documents has many significant applications in natural language processing (NLP) and information retrieval (IR) such as plagiarism detection \cite{barron2013plagiarism}, query ranking \cite{figueroa2013learning}, duplicate question detection \cite{Wang:2017:BMM:3171837.3171865, bogdanova2015detecting}, web searching \cite{wang2013paraphrasing}, and automatic question answering \cite{fader2013paraphrase}.
	

	
	
	
	
	Paraphrase detection is a binary classification problem in which pairs of texts are labeled as either positive (paraphrase) or negative (non-paraphrase). In this setting, pairs of texts are mapped into a fixed-dimensional feature-space, where a standard classifier is learned. Feature maps based on lexical, syntactic and semantic similarities in conjunction with SVM are proposed in \cite{dey2016paraphrase, eyecioglu2015twitter}.  More recently, it has been demonstrated that for short text, deep learning-based pairs representations and classification yield better accuracy~\cite{agarwal2018deep}.
	
	

	
	

	
	Many deep learning-based schemes employ one or two Convolutional Neural Network (CNN) or Long Short Term Memory (LSTM) based models to learn features and make predictions on clean texts \cite{Wang:2017:BMM:3171837.3171865, hu2014convolutional, tomar2017neural}, while a recent model also incorporates linguistic features to detect paraphrases in both clean and noisy short texts \cite{agarwal2018deep}. For many NLP tasks involving short texts, it has been shown that developing wider models can yield significant gains \cite{reimers2017reporting}.
	
	While deep models produce richer representations, they require large amounts of training data for a robust paraphrase detection system ~\cite{dey2016paraphrase}. Thus, for small datasets, such as Microsoft Research Paraphrase (MSRP) corpus and SemEval-2015 Twitter paraphrase dataset (SemEval), handcrafted features and SVM classifier have been widely used~\cite{dey2016paraphrase, ji2013discriminative}.
	Labeling pairs of documents in a human-based computation setting (e.g. crowd-sourcing) is costly \cite{xu2015semeval}. Therefore, \cite{agarwal2018deep} and \cite{tomar2017neural} add to the training set each labeled pair also in the reversed order. However, this simple data augmentation strategy can be extended in a systematic manner by relying upon set and graph theory. For instance, consider four documents: ($a$) \textit{How can I lose weight quickly?} ($b$) \textit{How can I lose weight fast?} ($c$) \textit{What are the ways to lose weight as soon as possible?} ($d$) \textit{Will Trump win US elections?}. If in the annotated corpus, documents $a$ and $b$ and documents $b$ and $c$ are marked as paraphrases, then by transitive extension, documents $a$ and $c$ can also be considered as paraphrases. Similarly, if documents $a$ and $b$ are labeled as paraphrase, while documents $b$ and $d$ are labeled as non-paraphrase, then a new non-paraphrase pair based on document $a$ and $d$ can be inferred reliably. Such a strategy can be used to generate additional annotations in a sound and cost-effective manner, and potentially enhance the performance of deep learning models for paraphrase detection. 
	
	In this paper, we propose a data augmentation strategy for generating additional paraphrase and non-paraphrase annotations reliably from existing annotations. We consider notions of paraphrases and non-paraphrases as binary relations over the set of documents. Representing the binary relation induced by the paraphrase labels as an undirected graph and performing transitive closure on this graph, we include additional paraphrase annotation in the training set. Similarly, by comparing paraphrase and non-paraphrase annotations we infer additional non-paraphrase annotations for inclusion in the training corpus. Our strategy involves several steps and a parameter through which the data augmentation can be tuned for enhanced paraphrase detection.


We also present a robust multi-cascaded deep learning model for paraphrase detection in short texts. Our model utilizes three independent CNN and LSTM (with and without soft attention) cascades for feature learning in a supervised manner. We also employ a number of additional linguistic features after corpus-specific text preprocessing. All these features are fed into a discriminator network for final classification. 
	 
	 
To show effectiveness of our approach we evaluate the data augmentation and deep model on three benchmark short text datasets (MSRP and Quora (clean), and SemEval (noisy)). We also perform extensive comparisons with the state-of-the-art methods. 
We make the following key contributions in this work: 
\begin{itemize}
	\item We present an efficient strategy for augmenting existing paraphrase and non-paraphrase annotations in a consistent manner. This strategy generates additional annotations and enhances the performance of the data-hungry deep learning models. 
	\item We develop a multi-cascaded learning model for robust paraphrase detection in both clean and noisy texts. This model incorporates multiple learned and linguistic features in a wide and deep discriminator network for paraphrase detection. 
	\item We address both clean and noisy texts in our presentation and show that the proposed model matches current best performances on benchmark datasets of both types.
	\item We analyze the impact of various data augmentation steps and different components of the multi-cascaded model on paraphrase detection performance. 
\end{itemize}

The rest of the paper is organized as follows: We discuss the related work in paraphrase detection and data augmentation in Section \ref{sec:relatedWork}. We present our data augmentation strategy in Section \ref{sec:dataAugmentation}. Our multi-cascaded model for paraphrase detection is presented in Section \ref{sec:ProposedModel}. Section \ref{sec:ExperimentalEvaluation} outlines our experimental evaluation setup including discussion of data augmentation. We present and discuss the results of our approach in Section \ref{sec:Results}. Finally, we present our concluding remarks in Section \ref{sec:Conclusion}.



	\section{Related Work} \label{sec:relatedWork}
	
	Automatic paraphrase detection has been widely studied in the NLP and IR communities. The problem is posed as classification of pairs of text into one of paraphrase and non-paraphrase classes. In this setting, first, the pairs of texts are represented as fixed-length vectors. This representation tends to be sparse due to short length of texts~\cite{rashid2019fuzzy}. Thus, this representation must be efficiently computable and should preserve as much contextual information as possible. Standard classifiers, such as SVM, are then learned in this representation scheme for detecting paraphrases. 
 Short text can be clean (i.e., that follows proper grammar and formal diction like news headlines) or noisy (i.e., having informal verbiage and spelling variations like tweets). An abundance of work has been done on clean text paraphrase detection. A weighted term frequency approach for text pair representation along with $n$-gram overlap features between two texts is proposed in~\cite{ji2013discriminative} to train SVM classifier. In \cite{mohammad2017paraphrase}, lexical (Parts of Speech (POS) overlap, minimum edit distance, text alignment) and semantic (Named-entity overlap, topic modeling) features between a pair of input text are used to train support vector regressor for paraphrase identification of news tweets in Arabic language. A probabilistic model that relies upon the similarity between syntactic trees of two input documents is proposed in~\cite{das2009paraphrase}. The authors in \cite{oliva2011symss} devise a method for computing semantic similarity based on the lexical database. The model depends on WordNet meanings and syntactic roles among words in two documents.

Several studies have been carried out that utilize deep learning architectures for paraphrase detection in clean short text. A recursive auto-encoder for reliably understanding the context of texts and performing paraphrase detection is proposed in \cite{socher2011dynamic}. This architecture forms a recursive tree and performs dynamic pooling to convert the input into fixed-sized representations. However, making a tree requires parsing hence this approach is less scalable. In \cite{hu2014convolutional}, patterns learned on a pair of text through CNN are matched at different levels of abstraction, introducing explicit interaction between the two documents during the learning process. In \cite{el2015boosting}, five lexical metrics are used for reliable semantic similarity detection, where abductive networks are employed to get a composite metric that is used for classification. A method of decomposing text pair to similar and dissimilar components is proposed in \cite{wang2016sentence}. A CNN is then trained to convert these components into a fixed dimension vector and classification. In \cite{yin2016abcnn}, three attention schemes are used in CNN to form interdependent document representations.  Many neural network models are proposed to match documents from multiple levels of granularity. A multi-perspective matching model ({BiMPM}) is proposed in \cite{Wang:2017:BMM:3171837.3171865}, that uses character-based LSTM to learn word representations and a bi-directional LSTM for document representation for each text. After that, it performs four types of matching in ``matching-layers" and finally, all these representations are aggregated by an additional bi-directional LSTM for paraphrase detection. One extension of this work is the neural paraphrase detection model based on a self-attended feed-forward network with pretrained embeddings on a huge corpus of another paraphrase data \cite{tomar2017neural}. It is shown that this approach outperforms BiMPM model in terms of testing accuracy with fewer parameters.

As compared to clean short text, less work has been done for noisy short text paraphrase detection. In \cite{xu2014extracting}, string-based features (whether the two words, their stemmed forms, and their normalized forms are the same, similar or dissimilar), common POS, and common topic (word's association with a particular topic) between a pair of text are used as features and a novel multi-instance learning paraphrase model (MultiP) is proposed. Simple lexical features based on word and character $n$-gram overlaps between two texts are constructed to train SVM classifier in \cite{eyecioglu2015twitter}. An approach in \cite{zhao2015ecnu} uses corpus-based (similarity between sum of word vectors of two sentences), syntactic, and sentiment polarity based features and train SVM classifier. While \cite{zarrella2015mitre} uses ensemble approach based on seven models using word embeddings. In \cite{dey2016paraphrase}, a set of lexical (character and word level $n$-grams), syntactic (words with matching POS tag, same verbs), semantic (adjective overlap), and pragmatic (subjective/objective agreement) features are identified between pair of input text, along with extensive preprocessing (spelling correction, stemming, stopwords removal, synonymous replacement). Although these features perform well on noisy short text, it is shown that they fail to get good predictive performance on clean text corpus of Microsoft Research Paraphrase (MSRP).
    
	All previously reported approaches were focused either on clean text or noisy text. A recent study focuses to develop a single deep learning model that performs well on both clean and noisy short text paraphrase identification \cite{agarwal2018deep}. A CNN and LSTM-based approach is adopted to learn sentence representations, while a feedforward network is used for classification. It also utilizes hand-crafted linguistic features, which improves paraphrase detection accuracy. Our approach follows this trend and we focus on a single model for paraphrase identification in both clean and noisy datasets. We also use linguistic features but our set of these features is different than~\cite{agarwal2018deep}.
	
	Deep learning approaches need a large-scale annotated dataset for developing a robust model. For instance, the AskUbuntu dataset \cite{dos2015learning} contains very few annotations, thus limiting the generalization performance of the model \cite{tomar2017neural}. The ability to augment the data with additional sound annotations without requiring human intervention can improve the performance of deep models \cite{agarwal2018deep, tomar2017neural}. Such data augmentation has been shown to be fruitful for data analytics when only a piece of limited ordinal information about the pairwise distance between objects is provided \cite{kleindessner2017lens, NIPS20177257, heikinheimo2013crowd}. Data augmentation has also been shown to be prolific in image classification \cite{wang2017effectiveness}. Here, standard image processing such as cropping, rotation, and object translation is done to generate additional image samples. To the best of our knowledge, a systematic procedure for augmenting paired data without relying upon the object’s content has not been presented earlier.

	\section{Data Augmentation for Paraphrase Detection} \label{sec:dataAugmentation}
	
	We start the presentation of our enhanced approach for paraphrase detection by discussing the proposed data augmentation strategy. Paraphrase annotation is costly and time-consuming while deep learning approaches demand a large corpus of annotated paraphrases. To address this problem, we develop strategies for generating additional annotations efficiently in a sound manner. We rely upon set theory and graph theory to model the paraphrase annotation problem and present an algorithm for generating additional data for training. 
	
	Let $\mc{D} = \{d_1, d_2, \ldots, d_{|\mc{D}|}\}$ be the set of documents in the annotated corpus. The corpus contains annotations for paraphrases and non-paraphrases. The triplet $(d_i, d_j, 1)$ indicates that documents $d_i \in \mc{D}$ and $d_j \in \mc{D}$ are considered paraphrases, and the triplet $(d_i, d_j, 0)$ denotes that documents $d_i \in \mc{D}$ and $d_j \in \mc{D}$ are considered non-paraphrases. Let $N_p$ and $N_{np}$ denote the numbers of paraphrase and non-paraphrase annotations in the corpus, and $N = N_p + N_{np}$ be the total number of annotations in the corpus. Note that in practice, only a fraction of the pairs of documents in $\mc{D}$ will be annotated in the corpus, i.e., $N \ll |\mc{D}|^2$.  
	
	The information contained in the annotated corpus can also be represented as a graph over the vertex set in $\mc{D}$. Each triplet corresponds to an edge in the graph with its label ($1$ or $0$) indicating whether the two documents are considered paraphrases or not. For example, the triplet $(d_i, d_j, 1)$ is represented by an edge between vertex $d_i \in \mc{D}$ and vertex $d_j \in \mc{D}$ with label $1$. We assume that each edge can have a single label only, i.e., there are no conflicts in the annotated corpus whereby the same pairs of documents are labeled as both $1$ and $0$. If such conflicts do exist, they are removed from the corpus.

	\subsection{Generating Additional Paraphrase Annotations} 
	\label{sc:gen_paraphrases}
	The notion of pairs of documents in $\mc{D}$ being considered paraphrases can be captured by the notion of binary relation in set theory. Let $\mc{R}_p \subset= (\mc{D} \times \mc{D})$ define the binary relation over $\mc{D}$ such that $\forall{i} \forall{j} (d_i, d_j) \in \mc{R}_p$   implies that $d_i$ is a paraphrase of $d_j$. In general, the following two properties hold for $\mc{R}_p$: 
	\begin{enumerate}
		\item $\mc{R}_p$ is reflexive, i.e., $\forall{i}, (d_i, d_i) \in \mc{R}_p$. 
		\item $\mc{R}_p$ is symmetric, i.e., $\forall{(i,j)}, (d_i, d_j) \in \mc{R}_p \implies (d_j, d_i) \in \mc{R}_p$. 
	\end{enumerate} 
	
	The notion of paraphrasing is not defined precisely in linguistics. The boundary between paraphrases and non-paraphrases can lie on the continuum between (strong) paraphrases on one end and (strong) non-paraphrases on the other \cite{vila2014paraphrase}. For clean texts (e.g., news headlines), we may approximate the notion of paraphrase by the notion of semantic duplicate, i.e., $(d_i, d_j, 1)$ implies that documents $d_i$ and $d_j$ are considered duplicates semantically. In set theory, this corresponds to the equivalence relation. The equivalence relation $\mc{R}_e$ over $\mc{D}$ possesses the following property in addition to properties 1 and 2 listed above:

	\begin{enumerate}
		\setcounter{enumi}{2}
		
		\item $\mc{R}_e$ is transitive, i.e., $\forall{(i  \neq j \neq k)}, [(d_i, d_j) \in \mc{R}_e  \land  (d_j, d_k) \in \mc{R}_e] \implies (d_i, d_k) \in \mc{R}_e$
	\end{enumerate}
	
	Transitivity can be a strong property when applied to the notion of paraphrases, especially for noisy text. Therefore, we consider $\mc{R}_p$ to include transitive extensions of the direct relation $\mc{R}$ to a pre-selected order $K\geq 1$. That is, $\mc{R}_p = \mc{R}_p \cup \mc{R}  \cup \mc{R}^1 \cup \cdots \cup \mc{R}^K$ where $\mc{R}$ denotes the relation that two documents are paraphrases due to a direct relationship between them and relation $\mc{R}^K$ indicates that two documents are considered paraphrases because they are $K \geq 1$ intermediate documents relating them ($\mc{R}^1$ is the transitive extension of $\mc{R}$, and $\mc{R}^K$ is the transitive extension to order $K$ of $\mc{R}$). If $K = *$ (i.e., maximum order extension is done) then we achieve a transitive closure of $\mc{R}$. 
	
	Now, consider the graph on the vertex set $\mc{D}$ induced by edges labeled with 1, i.e., pairs considered to be paraphrases in the annotated corpus. These pairs do not necessarily induce the relation $\mc{R}_p$ on $\mc{D}$. We therefore add more pairs to transform it into the desired relation. More formally:
	
	\begin{enumerate}
		\item For each $d_i \in \mc{D}$, we add $(d_i, d_i, 1)$ in the corpus, i.e., we declare each $d_i$ a paraphrase of itself. We call this step generation by reflexivity.  
		\item For each $(d_i, d_j, 1)$, in the corpus, we add $(d_j, d_i, 1)$ in the corpus, i.e., we consider $d_j$ and $d_i$ to be paraphrases of each other. We call this step generation by symmetry. 
		\item For every chain of annotations $(d_i, d_{j_1}, 1), (d_{j_1}, d_{j_2}, 1), \ldots , (d_{j_{K-1}}, d_{j_K}, 1), (d_{j_K}, d_k, 1)$ starting at $d_i$ and ending at $d_k$ with at most $K$ intermediate documents in the annotated corpus, we add $(d_i, d_k, 1)$ and $(d_k, d_i, 1)$ in the corpus. Thus, we consider $d_i$ and $d_k$ to be paraphrases of each other if these documents are connected by at most $K$ intermediate documents. We call this step generation by paraphrase extension.  
	\end{enumerate}
	In graph terminology, step 1 corresponds to adding self-loops on each vertex with label 1 while in step 2, we ignore the direction on all edges by considering the graph as an undirected graph. Step 3 corresponds to performing a transitive extension on the undirected graph induced by edges labeled with 1. Every vertex needs to be made adjacent to all vertices that are reachable from it in $\leq K$ hops. This can be done with a single BFS (breadth first search) or DFS (depth first search) on the graph.
	
	Note that transitive extension transforms the graph into a collection of cliques (fully connected vertices) where each clique is a paraphrase class representing a unique concept.

	\subsection{Generating Additional non-Paraphrase Annotations }
	\label{sc:gen_non_paraphrases}
	Let $\mc{R}_{np}$ denote the binary relation that two documents in $\mc{D}$ are considered non-paraphrases. By definition, this relation is irreflexive, i.e., $\forall{i}, (d_i, d_i) \not\in \mc{R}_{np}$. Obviously, a document cannot be a non-paraphrase of itself. The relation $\mc{R}_{np}$ is symmetric, i.e., $\forall{i, j}, (d_i, d_j) \in \mc{R}_{np} \implies (d_j, d_i) \in \mc{R}_{np}$. Transitivity does not hold for relation $\mc{R}_{np}$; if $(d_i, d_j) \in \mc{R}_{np}$ and $(d_j, d_k) \in \mc{R}_{np}$, then we cannot say for sure that $d_i$ is not a paraphrase of $d_k$. 
	
	Additional non-paraphrase annotations can also be inferred by comparing them with paraphrase annotations. For example, if $(d_i, d_j, 0)$ and $(d_j, d_k, 1)$ (i.e., $d_i$ and $d_j$ are non-paraphrases, while $d_j$ and $d_k$ are paraphrases), then $(d_i, d_k, 0)$ must also be true (i.e., $d_i$ and $d_k$ are non-paraphrases). Of course, if $d_i$ and/or $d_k$ lie within two different paraphrase classes, then all pairs of documents in the classes will be considered non-paraphrases. These relations are also included in $\mc{R}_{np}$.    
	
	Now, consider the graph on vertex set $\mc{D}$, induced by edges labeled with 0, i.e., non-paraphrases in the annotated corpus. These pairs do not necessarily induce the relation $\mc{R}_{np}$ on $\mc{D}$. We, therefore, add more pairs to transform it into the desired relation as follows: 
	\begin{enumerate}
		\item For each $(d_i, d_j, 0)$, in the corpus, we add $(d_j, d_i, 0)$ in the corpus, i.e., we consider $d_j$ and $d_i$ to be non-paraphrases as well. We call this step generation by (non-paraphrase) symmetry  .
		
		\item For every $(d_i, d_j, 0)$ in the corpus, let $\mc{C}$ and $\mc{C}^\prime$ be the cliques (paraphrase classes) containing $d_i$ and $d_j$, respectively, we add $(d_m, d_n, 0)$ and $(d_n, d_m, 0)$  for each $d_m \in \mc{C}$ and $d_n \in \mc{C}^\prime$ to the corpus. We call this step generation by non-paraphrase extension.  
	\end{enumerate}
	
	In terms of graph, the second step corresponds to making a complete bipartite graph between the vertex set of $\mc{C}$ and that of $\mc{C}^\prime$.
	
	\subsection{Conflicts and Errors in Annotations }
	\label{sc:conflicts_errors} 
	Using the principled strategy outlined earlier, conflicts and errors in annotations can be identified and potentially fixed. A conflict occurs when a pair of documents is found to be paraphrase and non-paraphrase either in the original annotated corpus or arises during augmentation. Based on our data augmentation strategy described above, the following conflicts can arise: (1) In the original annotated corpus, a pair of documents is labeled as both paraphrase and non-paraphrase. (2) Erroneous annotations can be generated during our data augmentation strategy in the following two cases: (a) when generating additional paraphrases by a transitive extension (b) when generating additional non-paraphrases by non-paraphrase extension. A detailed analysis of conflicts and errors and their resolution is beyond the scope of this paper in which we focus on data augmentation and its impact on paraphrase detection. Nonetheless, we believe that this is a fruitful area for future research. 
	
	In this work, we resolve the first type of conflict by removing the conflicting annotations and the second type of conflict by removing the conflicting non-paraphrase annotation and retaining the generated paraphrase annotation. We control the number of conflicts and errors by varying $K$ and selecting/dropping specific generation steps. And we evaluate these variations by their performances on paraphrase detection.

	\begin{algorithm}[tp!]
		\caption{: Data augmentation strategy}
		\label{algo:paraphrase}
		\begin{algorithmic}[1]
			\State \textbf{Input: }$\mc{D}$ (documents),  $\mc{A}$ (original corpus), $K$ (extension order)
			\State \textbf{Output:} $\mc{\bar{A}}$ (augmented corpus)
			\LineComment{Remove conflicting annotations}
			\ForAll{$[(d_i, d_j, 0) \in \mc{A} \lor (d_j, d_i, 0) \in \mc{A}] \land [(d_i, d_j, 1) \in \mc{A} \lor (d_j, d_i, 1) \in \mc{A}]$}
			\State $\mc{A} \leftarrow \mc{A} \setminus \{(d_i, d_j, 0), (d_j, d_i, 0), (d_i, d_j, 1), (d_j, d_i, 1)\}$
			\EndFor
			\State $\mc{\bar{A}} \leftarrow \mc{A}$ \LineComment{Paraphrase augmentation} \LineComment{Step P1: by reflexivity}
			\ForAll{$d_i \in \mc{D}$}
			\State $\mc{\bar{A}} \leftarrow \mc{\bar{A}} \cup (d_i, d_i, 1)$
			\EndFor 
			\LineComment{Step P2: by symmetry}
			\ForAll{$(d_i, d_j, 1) \in \mc{A}$}
			\State $\mc{\bar{A}} \leftarrow \mc{\bar{A}} \cup (d_j, d_i, 1)$
			\EndFor
			\LineComment{Step P3: by transitive extension}
			\State $\mc{A} \leftarrow \mc{\bar{A}}$
			\For{($ n \leftarrow 1 \to K$)}
			\ForAll{$(d_i, d_{j_1}, 1) \in \mc{A} \land \cdots \land (d_{j_n}, d_k, 1) \in \mc{A}$}
			\If{$(d_i, d_k, 0) \in \mc{\bar{A}} \lor (d_k, d_i, 0) \in \mc{\bar{A}}$}
			\State $\mc{\bar{A}} \leftarrow \mc{\bar{A}} \setminus \{(d_i, d_k, 0), (d_k, d_i, 0)\{$
			\EndIf
			\State $\mc{\bar{A}} \leftarrow \mc{\bar{A}} \cup \{(d_i, d_k, 1), (d_k, d_i, 1)\}$
			\EndFor
			\EndFor
			\LineComment{Non-paraphrase augmentation}
			\LineComment{Step NP1: by symmetry}
			\ForAll{$(d_i, d_j, 0) \in \mc{\bar{A}}$}
			\State $\mc{\bar{A}} \leftarrow \mc{\bar{A}} \cup (d_j, d_i, 0)$
			\EndFor
			\LineComment{Step NP2: by non-paraphrase extension}
			\ForAll{$(d_i, d_j, 0) \in \mc{\bar{A}} \land (d_j, d_k, 1) \in \mc{\bar{A}}$}
			\State $\mc{\bar{A}} \leftarrow \mc{\bar{A}} \cup \{(d_i, d_k, 0), (d_k, d_i, 0)\}$ 
			\EndFor 
		\end{algorithmic}
	\end{algorithm}	
	
	\subsection{Algorithm}
	Algorithm 1 outlines our proposed strategy for augmenting data for enhanced paraphrase detection. The algorithm takes as input the set of documents $\mc{D}$, the (original) annotated corpus or dataset $\mc{A}$, and the parameter $K$ (extension order) and it outputs the augmented annotated corpus or dataset $\mc{\bar{A}}$. After removing conflicts in the dataset (lines $4-5$), the algorithm proceeds with generating additional paraphrase annotations (lines $6-15$) followed by generating additional non-paraphrase annotations (lines $16-19$). Generating paraphrase annotations involve $3$ steps i.e., P1 (lines $6-7$), P2 (lines $8-9$), and P3 (lines $10-15$), while generating non-paraphrase annotations consists of two steps i.e., NP1 (lines $16-17$) and NP2 (lines $18-19$). It is worth noting that step P1 can be performed at any sequence while the other steps must follow the given sequence. In our experiments, we perform P1 after P2 and P3. Note that step P1 means additionally generated paraphrase pairs would be equal to the number of unique texts in the training data (minus number of pairs that were already part of the original annotations).
	
	The worst-case computational complexity of the algorithm is defined by step P3. If $Z$ is the size of the largest paraphrase class (max. clique size or max. node degree in graph terminology), then the computational complexity of the algorithm is $O(|\mc{D}| Z K)$. Note that in practice both $Z$ and $K$ will be much less than $|\mc{D}|$.

	
	\section{Multi-cascaded Deep Model for Paraphrase Detection} \label{sec:ProposedModel}
	In this section, we present our multi-cascaded deep learning model for enhanced paraphrase detection. While deep models have been popularly used in recent years for paraphrase detection, they are typically tailored to either clean text (e.g., news headlines) or noisy text (e.g., tweets). Similarly, most employ a single model for feature learning and discrimination, and some do not utilize linguistic features in their models. In order to benefit from previous insights and to produce robust paraphrase detection for both clean and noisy texts, we propose three independent feature learners and a discriminator model that can consider both learned and linguistic features for paraphrase detection. 
	\begin{figure}[b!]
		\begin{center}

			\includegraphics[page=1, scale=0.55]{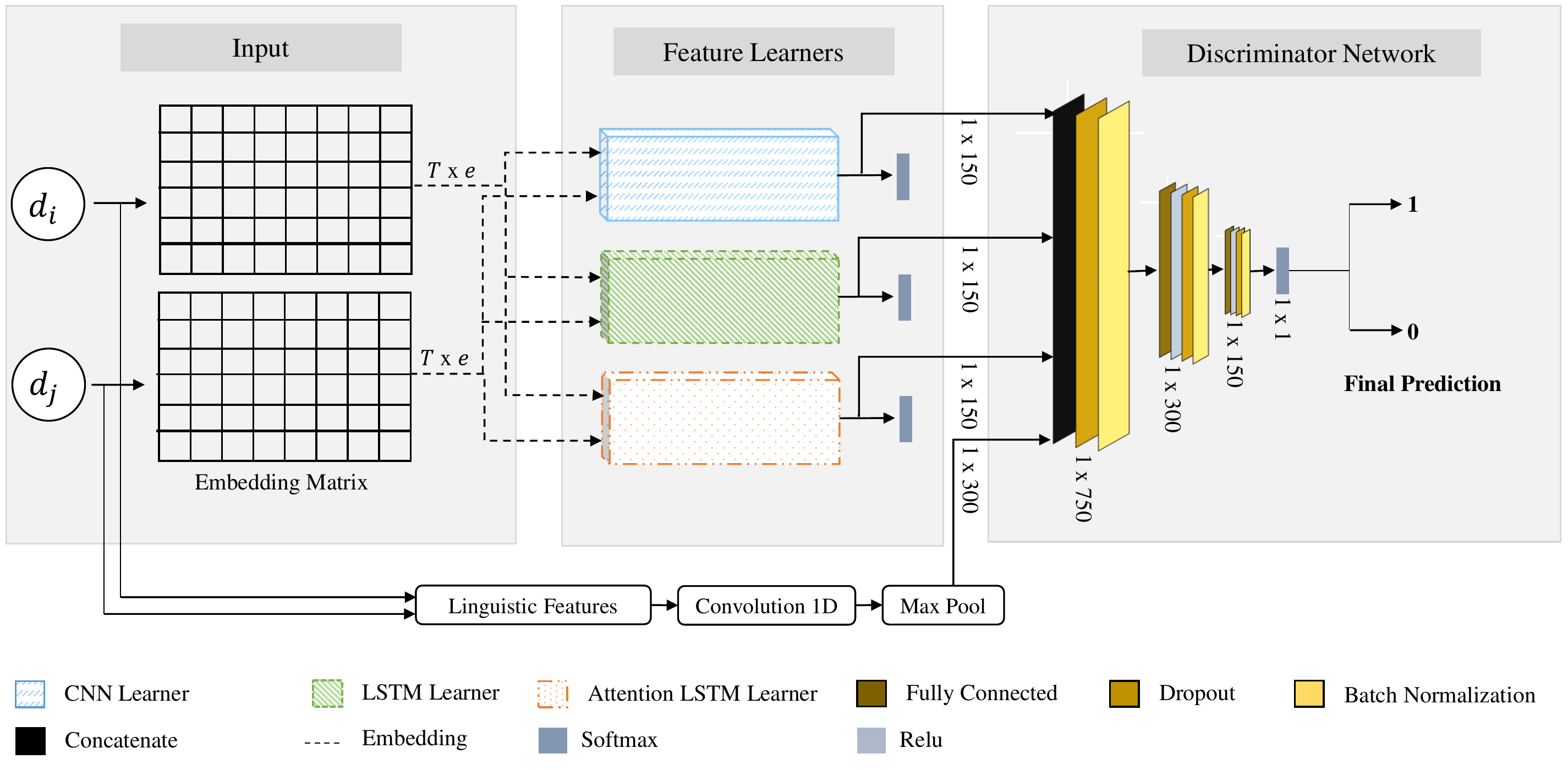}
			\caption{Multi-cascaded model for enhanced paraphrase detection (Figure best seen in color)}
			\label{fig:architecture}
		\end{center}
	\end{figure}

	Figure \ref{fig:architecture} shows the architecture of our multi-cascaded model for paraphrase detection. We employ three feature learners that are trained to distinguish between paraphrases and non-paraphrases independently (in parallel). The features from these models (one layer before the output layer) are subsequently fed into a discriminator network together with any additional linguistic features to make the final prediction.  
	
	Our model takes as input a pair of documents $d_i$ and $d_j$ and outputs the label 1 if the documents are considered paraphrases and the label 0 if the documents are considered non-paraphrases. Let $d_i = <w_{1}^i, w_{2}^i, \ldots, w_{T_i}^i>$ be the sequence of words in document $d_i$. We represent each word in the sequence by an $e$ dimensional fixed-length vector (word embedding). For a given paraphrase detection problem, we empirically decide length $T$ of each document to use such that longer documents are truncated and shorter ones are padded with zero vectors. This decision is made to ensure compatible length vectors for all document pairs across different models. As discussed in Section \ref{sec:Results}, we select an appropriate length $T$ for each dataset based on the distribution of lengths of documents in the dataset. We experiment with several linguistic features (syntactic and lexical) for both clean and noisy text paraphrase detection. The details of these features are given in Section \ref{subsec:nlpFeat}.   
	
	The details of the feature learners and the discriminator network are given in the following subsections.  
	\begin{figure}[b!]
		\begin{center}
			\includegraphics[page=1, scale=0.55]{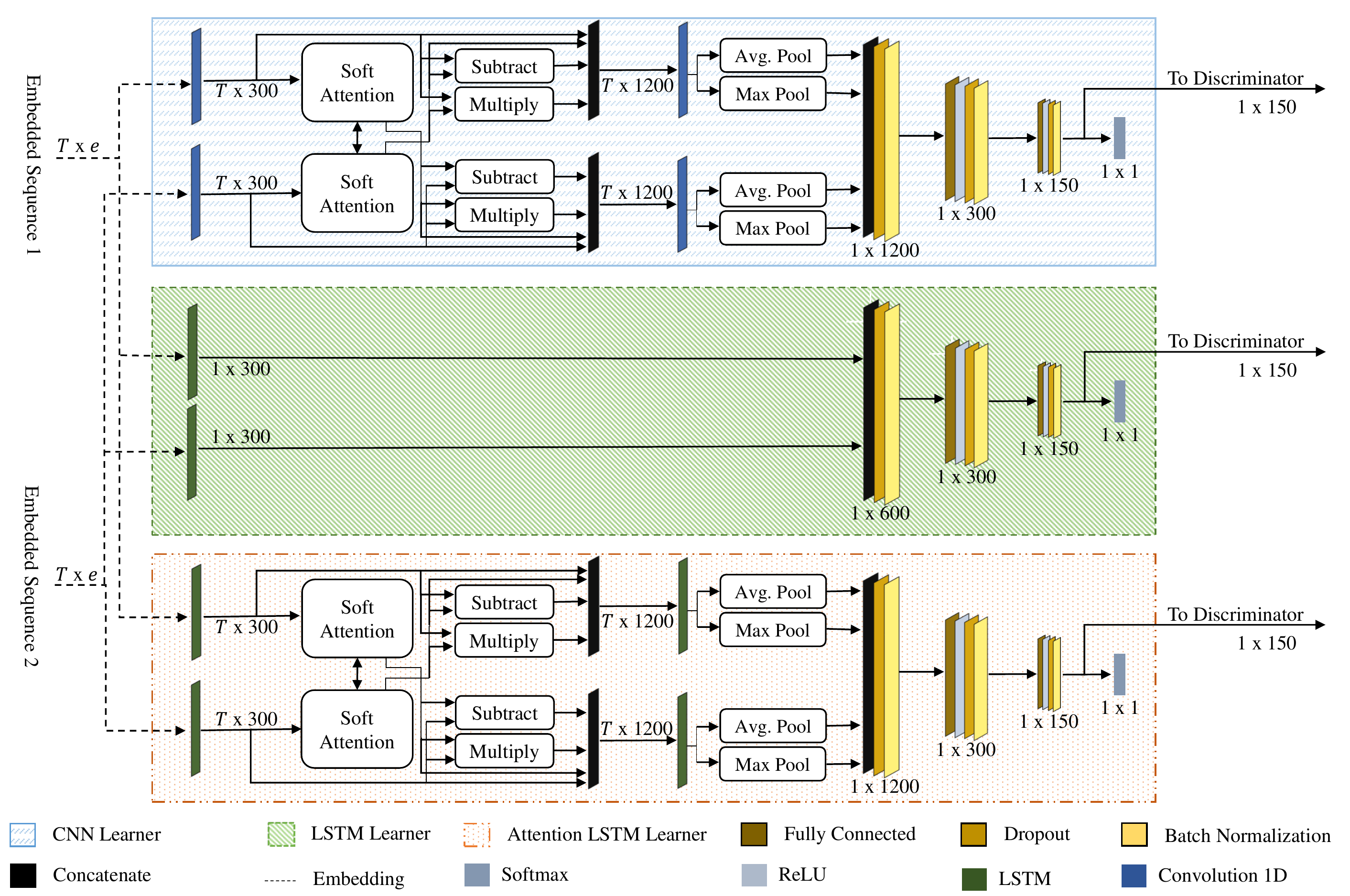}
			\caption{Architecture and dimensions of output for each cascade in our multi-cascaded model (Figure best seen in color) }
			\label{fig:cascades}
		\end{center}
	\end{figure}

	\subsection{Feature Learners}
	
	We employ three independent cascades to learn contextual features for paraphrase detection. Each cascade focuses on a different sequential learning model to extract features from different perspectives. Each cascade takes as input the pair of documents to be classified as paraphrase or non-paraphrase, and each is trained independently on the annotated corpus. The details of all three cascades are discussed in the following paragraphs. 
	
	The first cascade is based on CNN with soft-attention (Figure \ref{fig:cascades}). The first layer has 300 CNN filters with kernel size of 1. Subsequently, soft-attention is applied to highlights words in documents $d_i$ and $d_j$ that are more important to achieve correct prediction. The result of soft-attention for document $t_i$ is then subtracted and multiplied with that of document $t_j$ to learn semantic contrariety. The next layer concatenates the output of CNN layer, soft-attention output, subtracted output, and multiplied output for both documents (one layer concatenates these 4 outputs for $d_i$ and other for $d_j$ independent of each other). This output is forwarded to another CNN layer containing 300 filters with kernel size of 2. The purpose of this CNN layer is to learn bi-gram feature representation. After this representation, global max-pooling and global average-pooling is performed to obtain most important bi-gram and average of all bi-grams respectively for both documents. These two representations are then concatenated to make a single vector of both documents (up till now, both documents were being treated separately so two streams of same functions were being produced). This outputs a vector of length 1,200. After this concatenation, a dropout and batch normalization layer is deployed to avoid any feature co-adaptation. Then this representation is forwarded to a fully connected layer with 300 units with ReLU activation followed by drop out and batch normalization. Finally, this representation is squashed into a 150-dimensional vector by another fully connected layer. This cascade is learned to detect paraphrases and non-paraphrases, and the learned 150-dimensional representation is forwarded to discriminator network for final classification.
	
	The second cascade utilizes LSTM to encode long-term dependencies in the documents \cite{wang-etal-2016-combination} (Figure \ref{fig:cascades}). This cascade learns contextual representations without any attention or semantic contrariety. As such, it comprises of a single LSTM layer with 300 units followed by fully connected hidden layers with 300 and 150 units each and an output layer. Again, the 150-dimensional representation serves as an input to the discriminator network for final classification.  
	
	The third cascade is based on LSTM with soft-attention (Figure \ref{fig:cascades}). This cascade is similar to the first cascade but uses LSTM units instead of CNN filters. As such, it does not learn bi-gram feature representation but instead learns long-term dependencies with important sequences highlighted by attention mechanism. The 150-dimensional vector obtained before the output layer is used as an input to the discriminator network for final classification. 
	
	We empirically decide to use 300 CNN filters, 300 LSTM units, 300 units for first fully-connected layer, and 150 units for second fully-connected layer in all cascades. In all feature learning cascades, weights of CNN filters as well as weights of LSTM layers are shared among $d_i$ and $d_j$. Weight sharing reduces the number of parameters required and both documents are converted to deep representations in same embedding space. We use ReLU activation for every layer except last prediction layer, which uses the softmax function. We train the model using categorical cross entropy loss. Dropout and batch normalization is used after every fully-connected layer.
	
	\subsection{Discriminator Network}
	
	Figure \ref{fig:architecture} shows the discriminator network utilized to make the final prediction. For a given pair of documents $d_i$ and $d_j$, this network takes as input the features from all three cascades plus any linguistic features with dropout and batch normalization. Two fully-connected hidden layers are then used with 300 and 150 units, respectively, where each is followed by dropout and batch normalization layer. The activation function for both fully-connected layer is set to ReLU. On final layer softmax activation function is used with categorical cross-entropy as loss. This network outputs final prediction for documents $d_i$ and $d_j$ as either paraphrase or non-paraphrase. We also do early-stopping if validation accuracy of discriminator is not improved for $10$ epochs. A checkpoint of the model is created after the training epoch at which validation accuracy is improved.

	
	\section{Experimental Setup} \label{sec:ExperimentalEvaluation}
	In this section, we describe the settings for the experimental evaluation of our enhanced paraphrase detection model. We discuss the key characteristics of the three datasets used in our evaluations before and after augmentation. We also present the parameter settings of the model, the text preprocessing performed, and the linguistic features used in our experiments.

	\subsection{Datasets and their Augmentation}
	
	We use three real-world datasets in our experimental evaluations. The Quora questions pairs dataset (Quora) \cite{Wang:2017:BMM:3171837.3171865} contains pairs of questions with their annotations (paraphrase or non-paraphrase)\footnote{https://github.com/zhiguowang/BiMPM}. This dataset is collected from questions posted on the Quora question-answering website. This is a clean text dataset. The Microsoft Research Paraphrase Corpus (MSRP) \cite{dolan2004msr} contains pairs of sentences together with their paraphrase/non-paraphrase annotation\footnote{https://www.microsoft.com/en-us/download/details.aspx?id=52398}. The sentences are extracted from news articles on the web. Thus, this is another example of a clean text dataset. The SemEval-2015 Twitter paraphrase dataset (SemEval) \cite{xu2014extracting} contains pairs of tweets in English with their paraphrase/non-paraphrase annotation \footnote{https://github.com/cocoxu/SemEval-PIT2015}. This is a noisy text dataset. The Quora and SemEval datasets have pre-defined training, development, and test sets available. The MSRP dataset has predefined training and test sets only. The statistics for each split of all the datasets are presented in Table~\ref{tab:datasetStats}. We describe the characteristics of the \emph{training} set of each dataset before and after augmentation in the following subsections.

	\begin{table}[t!]
		\centering
		\caption{Statistics for all datasets}
		\label{tab:datasetStats}
		
		\begin{tabular}{cccccc}
			\toprule
			
		Dataset & Split &Total Pairs & Paraphrase Pairs & Non-paraphrase Pairs & Debatable Pairs \\
			\midrule
		\multirow{3}{*} {Quora} & Train & $384,348$ & $139,306$ & $245,042$ & $-$ \\
			& Dev & $10,000$ & $5,000$ & $5,000$ & $-$ \\
			& Test & $10,000$ & $5,000$ & $5,000$ & $-$ \\
			\midrule
			
		\multirow{2}{*}	{MSRP} & Train & $4,076$ & $2,753$ & $1,323$ & $-$ \\
			& Test & $1,725$ & $1,147$ & $578$ & $-$ \\
			\midrule
			
			\multirow{3}{*} {SemEval} & Train & $13,063$ & $3,996$ & $7,534$ & $1,533$ \\
			& Dev & $4,727$ & $1,470$ & $2,672$ & $585$ \\
			& Test & $972$ & $175$ & $663$ & $134$ \\

			\bottomrule 
		\end{tabular}
		
	\end{table}
	
	\subsubsection{Quora Dataset}
	The Quora dataset contains annotations for 517,968 unique questions. We notice that there are 30 incorrect non-paraphrase annotations in the dataset. The texts in these 30 annotations are identical but they are marked as non-paraphrases. We remove these annotations from the dataset. 
	
	An analysis of this dataset reveals that questions have an average length of about $13$ words with a standard deviation of $6.7$ words and a maximum length of $272$ words.   Based on this analysis, we select $T = 40$ for this dataset. 
	
	\begin{table}[t!]
		\centering
		\caption{Numbers of paraphrase and non-paraphrase annotations in the datasets before and after augmentation }
		\label{tab:augPairs}
		
		\begin{tabular}{lccccccc}
			\toprule
			
			\multirow{2}{*}{Dataset} & \multicolumn{4}{c}{Paraphrases}  & \multicolumn{3}{c}{Non-Paraphrases}\\ 
			\cmidrule(lr){2-5} \cmidrule(l){6-8}
			& Original & P2 & P3 & P1& Original &NP1 & NP2  \\
			\midrule

			Quora & $139,306$ & $278,612$ & $447,378$ & $964,509$ & $245,042$ & $490,015$ & $655,219$  \\
			
			MSRP & $2,753$ & $5,506$ & $5,874$ & $13,688$ & $1,323$ & $2,646$ & $2,647$  \\
			
			SemEval & $3,996$ & $7,992$ & $94,722$ & $107,953$ & $7,534$ & $15,068$ & $23,205$  \\

			\bottomrule 
		\end{tabular}
		
	\end{table}

	Table \ref{tab:augPairs} shows the numbers of paraphrases and non-paraphrases in the Quora dataset before and after each step of augmentation. For this clean text dataset, we perform step P3 (paraphrase generation by transitive extension) using $K = *$, i.e., we generate by transitive closure. Applying step P2 (generation by symmetry) almost doubles the number of paraphrase annotations. Performing transitive closure (step P3) generates $168,766$ additional paraphrase annotations. Subsequently, generating paraphrase annotations by reflexivity (step P1) produces an additional $517,131$ paraphrase pairs, bringing the total of paraphrase annotations to $964,509$.
	
	For augmentation of non-paraphrase annotations, step NP1 (generation by symmetry) brings the total number of non-paraphrase pairs to $490,015$. Subsequently, performing step NP2 (generation by non-paraphrase extension) an additional $165,204$ non-paraphrase pairs are produced, bringing the total number of non-paraphrase annotations to $655,219$.

	After performing step P3 (generation by transitive closure), it is observed that there are $57,119$ unique paraphrase classes (cliques) corresponding to distinct concepts about which questions are being asked. Note that transitive closure converts the graph on paraphrase annotations into a disjoint collection of cliques. Therefore, the node degree after transitive closure is one less than the number of questions (nodes) in which that node lies. 
	
	\begin{table}[!bp]
		\centering
		\caption{Examples of errors detected during paraphrase augmentation using transitive extension on Quora dataset (1 = paraphrase, 0 = non-paraphrase)}
		\label{tab:conflicts}
		
		\begin{tabular}{lcc}
			\toprule
			Question Pair & Original Label & Generated Label \\ 
			\midrule
			How can I get free iTunes gift cards online? & & \\
			What 's the best way to legally get free iTunes gift cards?  & 0& 1\\
			\midrule
			What is the colour of the Sun? && \\
			What is the color of the sun? & 0& 1\\
			\midrule
			Is pro wrestling fake? && \\
			Wwe is real fight? & 0& 1\\
			
			\bottomrule
			
		\end{tabular}
		
	\end{table}
	As discussed in Section \ref{sc:conflicts_errors}, our data augmentation strategy can highlight conflicts and errors in the annotations. For example, a paraphrase annotation generated by transitive closure can be in conflict with an existing non-paraphrase annotation. Usually, a generated annotation is based on strong evidence of related annotations and may indicate an error in the existing annotations. We find $214$ such conflicts in the training set of the Quora dataset. The number of conflicts can also be considered as a measure of annotation quality. Table \ref{tab:conflicts} shows some examples of such conflicts in the Quora dataset. These results confirm that our data augmentation strategy works well to detect and distinguish unique concepts and generate new pairs along with their associated labels from existing annotations reliably.

	
	\subsubsection{MSRP Dataset} \label{MSRP_augmentation}
	
	The MSRP dataset is a much smaller dataset containing annotations for $7,814$ unique sentences. There are no conflicting annotations found in this dataset. We select $T=25$ based on exploratory analysis of the dataset which reveals that the average length of sentences is about $19$ words with a standard deviation of $5.1$ word and a maximum length of $31$ words.
	
	Table \ref{tab:augPairs} gives the numbers of paraphrases and non-paraphrases before and after augmentation in this dataset. There are only $4,076$ annotations in the original dataset out of which $2,753$ are for paraphrases and $1,323$ are for non-paraphrases. The number of paraphrases is doubled after performing step P2 (generation by symmetry), an additional $368$ paraphrase annotations are generated in step P3 (generation by transitive closure), and step P1 generates an additional $7,814$ paraphrase annotations. Step NP1 doubles the number of non-paraphrase annotations while step NP2 generates just one more non-paraphrase annotation. As seen from these results, transitive closure does not add many annotations implying that there are generally small paraphrase classes in this dataset.  
	
	In this dataset, there were no conflicts detected while employing the proposed augmentation strategy.

	\subsubsection{SemEval Dataset}
	
		\begin{table}[t!]
			\centering
		\caption{Numbers of paraphrase and non-paraphrase annotations after step P3 and step NP2, respectively, in SemEval dataset with different orders of transitive extension }
		\label{tab:kLevelAugPairs}
		
		\begin{tabular}{lccccccc}
			\toprule
			\multirow{2}{*}{Tr. extension} & \multicolumn{2}{c}{Paraphrases}  & Non-Paraphrases\\ 
			\cmidrule(lr){2-3} \cmidrule(l){4-4}
			& P3 & P1&  NP2  \\
			\midrule
			
			$K= 1$ &  $30,738$ & $43,969$ & $18,539$  \\
			$K = 2$ &  $59,538$ & $72,769$ & $23,490$  \\
			$K = 3$ & $90,042$ & $103,273$ & $24,175$  \\
			$K = *$ &  $94,722$ & $107,952$ &  $23,205$ \\

			\bottomrule
		\end{tabular}
		
	\end{table}
	The SemEval dataset has paraphrase and non-paraphrase annotations for $13,231$ unique tweets. The original dataset has $7,534$ non-paraphrase, $3,996$ paraphrase, and $1,533$ \emph{debatable} annotations \cite{xu2015semeval}. All existing studies on this dataset ignore the debatable annotations while reporting their results; therefore, we also choose to ignore these annotations in our experiments. No conflicting annotations are found in the dataset.  
	
	For this dataset, we choose $T=19$. This number corresponds to the maximum length of tweets in the dataset with the average length being around $8$ words. 
	
	Table \ref{tab:augPairs} presents the numbers of paraphrases and non-paraphrases in this dataset before and after each step of our data augmentation strategy. These numbers are obtained when transitive closure is performed during step P3, i.e., $K=*$. It is seen that the number of paraphrases jumps from $3,996$  to $107,953$ while the number of non-paraphrases increases from $7,534$ to $23,205$. A major increase in paraphrases occurs during step P3. This can be attributed to the following two reasons: 1) The SemEval is based on a dataset developed for the task of semantic similarity estimation. As such, the notions of paraphrase and non-paraphrase are not well separated which is then blurred by the process of transitive closure. 2) The SemEval dataset contains short and noisy text that makes annotation difficult and prone to errors. For such datasets, transitive closure can be a `blunt instrument’ during the data augmentation strategy. 
	
	Table \ref{tab:kLevelAugPairs} shows the numbers of paraphrases and non-paraphrases after steps P3, P1, and NP2 when transitive extension of orders $K=1$, $K=2$, $K=3$, and $K=*$ (transitive closure) is performed. The numbers for step P2 and NP1 are not shown in this table as they are identical to those given in Table \ref{tab:augPairs}. It is clear from this table that order $K$ controls the number of paraphrases that are generated during step P3. For example, when $K=1$ the number of paraphrases after step P3 is $30,738$ which is significantly lower than $94,722$  when $K=*$. Note that the number of non-paraphrases actually increases slightly as the order of transitive extension is reduced. This is due to the fact that more conflicts arise as $K$ is increased that are resolved by retaining the generated paraphrase annotation and discarding the conflicting non-paraphrase annotation, and fewer non-paraphrase annotations will produce fewer additional non-paraphrase annotations during non-paraphrase extension (step NP2). The number of conflicts during step P3 are $4$, $14$, $22$, and $23$ for $K=1$, $K=2$, $K=3$, and $K=*$, respectively.

	\subsection{Preprocessing} \label{subsec:preprocessing}
	Text preprocessing is an essential component of many NLP applications. However, in case of short text, common text preprocessing steps such as removing punctuations and stopwords can result in loss of information critical to the application \cite{eyecioglu2015twitter}. Therefore, we keep preprocessing to a minimum in our experiments. For SemEval dataset, which represents noisy texts, we perform lemmatization and correct commonly misspelled words such as \textit{dnt} to \textit{do not}. We use a predefined dictionary to map misspelled words to their standard forms. Preprocessed tweets are used while training our multi-cascaded model as well as to extract linguistic features. For the other two datasets, which represent clean texts, we perform preprocessing (stopword removal and lemmatization) only for computing linguistic features while the raw text is used in our multi-cascaded model. The details of the linguistic features are given in the next subsection.
	
	\subsection{Linguistic Features} \label{subsec:nlpFeat}
	We employ a set of NLP/linguistic features in our experiments as it has been shown that including linguistic features for paraphrase identification in short text can improve the performance of deep learning models \cite{agarwal2018deep}. We identify the following linguistic and statistical features to be used alongside learned features in our multi-cascaded model. 
	
	\begin{enumerate}
		\item $2$ features based on TF-IDF cosine similarity between documents $d_i$ and $d_j$, before and after removing stopwords and doing lemmatization.
		\item $4$ n-gram overlapping ratio features based on unigrams and bigrams that are common to a given document pair, divided by total number of n-grams in $d_i$ and $d_j$ respectively.
		\item $2$ features based on cosine similarity between ELMo \cite{peters2018deep} embeddings vectors of $d_i$ with $d_j$, before and after removing stopwords and doing lemmatization.
		\item $2$ features based on cosine similarity between Universal Encoder \footnote{https://tfhub.dev/google/universal-sentence-encoder/2} vectors of $d_i$ and $d_j$, before and after removing stopwords and doing lemmatization.
		\item $1$ feature based on count of unigrams that has same POS tag in $d_i$ and $d_j$.
		\item $6$ features based on length of intersection of character bigrams, trigrams and quadgrams of $d_i$ in $d_j$, before and after removing stopwords and doing lemmatization.
		\item $2$ features based on longest substring match in $d_i$ and $d_j$, before and after removing stopwords and doing lemmatization.
		\item 2 features based on longest subsequence match in $d_i$ and $d_j$, before and after removing stopwords and doing lemmatization.
		\item $3$ syntactic features based on number of Verbs, Nouns and Adjectives common in $d_i$ and $d_j$.
		\item $2$ Named-entity Recognition (NER) features in $d_i$ and $d_j$ based on number of same NER tags as well as numbers of same word-NER tuple.
	\end{enumerate}
	
	We use all linguistic features for clean text datasets and linguistic features $1-4$ only for the noisy text dataset. Linguistic features are passed through a single CNN layer with $300$ dimensions and then provided as input to discriminator network.

	\subsection{Hyper-parameters Tuning} \label{subsec:hyperParameterTuning}
	
	A number of hyper-parameters are required in our multi-cascaded model for paraphrase detection. We tune and select these hyper-parameters on the development sets of each dataset using a grid search. As MSRP dataset has train and test splits only, we hold-out $10\%$ of the training set as the development set for the purpose of hyper-parameters tuning. We decide the type of word embeddings among GloVe \footnote{https://nlp.stanford.edu/projects/glove/} \cite{pennington2014glove}, Word2Vec \cite{mikolov2013distributed}, and ELMo. For selecting an optimizer, we decide among nAdam, Adam, Adadelta, and SGD. We consider dropout rates $0.1$, $0.2$, $0.3$, $0.4$ and $0.5$ in our model. 
	
	\subsection{Performance Evaluation and Comparison}
	Since paraphrase detection is a binary classification problem, standard measures of performance can be used for evaluation. We report performances as percent accuracy, precision, recall, and F1-value on the test sets after training over the respective training set of the datasets. Each dataset has a fixed training and test sets. Therefore, results can be compared with previously reported results on the same datasets.

	\subsection{Implementation}
	
	We use networkX library in Python for graph analysis and data augmentation\footnote{https://networkx.github.io/}. For implementing and evaluating our multi-cascaded model we use Keras\footnote{https://keras.io/} as the front-end with TensorFlow\footnote{https://www.tensorflow.org/} on the backside. All model parameters or weights are initialized randomly, and to ensure reproducibility the random seed is fixed. 
	The code and implementation setting used in our experimental evaluation is available from the website\footnote{code and data will be made available after acceptance or on-demand during review process.}.  
	 
	\section{Results and Discussion} \label{sec:Results}
	In this section, we present and discuss the evaluation of our multi-cascaded model and data augmentations strategy in terms of paraphrase detection predictive performance. We first present results on each dataset with and without data augmentation and linguistic features. Subsequently, we discuss the performance of different components of our multi-cascaded model. Finally, we present the key takeaways from our experimental study. 
	
	\subsection{Quora Dataset}
	For this dataset, the hyperparameters of our model are tuned on the provided dev set. GloVe embeddings are selected as the best choice while among optimizers, nAdam with learning rate of $0.002$ is found to be optimal. Similarly, the dropout rate of $0.1$ is found to be optimal.

	\begin{table}[t!]
		\centering
		\caption{Paraphrase detection performance on Quora dataset }
		\label{tab:quoraResultsProposed}
		
		\resizebox{\textwidth}{!}{
		\begin{tabular}{lcccccccccc}
			\toprule
			\multirow{2}{*}{Augmentation} & \multicolumn{4}{c}{Learned Features}  & \multicolumn{4}{c}{Learned + Linguistic Features}\\ 
			\cmidrule(lr){2-5} \cmidrule(l){6-9} 
			& Accuracy & Precision & Recall & F1-score &  Accuracy & Precision & Recall & F1-score \\
			\midrule
			
			None & $89.4$ & $90.3$ & $88.4$ & $89.3$ & $89.8$ & $90.0$ & $89.6$ & $89.8$  \\
			P2, NP1 & $89.6$ & $90.6$ & $88.4$ & $89.5$ & $74.9$ & $67.6$ & $\textbf{95.5}$ & $79.2$\\
			P2, P3 & $89.9$ & $90.0$ & $89.7$ & $89.9$ & $\textbf{90.1}$ & $\textbf{91.2}$ & $88.7$ & $89.9$ \\
			P2, P3, P1 & $90.2$  & $90.3$ & $\textbf{90.1}$ & $\textbf{90.2}$ & $90.0$ & $90.2$ & $89.6$ & $89.9$\\
			P2, P3, NP1 & $90.0$  & $90.5$ & $89.5$ & $90.0$ & $90.0$ & $89.9$ & $90.1$ & $90.0$ \\
			P2, P3, P1, NP1 & $\textbf{90.3}$  & $\textbf{90.9}$ & $89.5$ & $\textbf{90.2}$ & $\textbf{90.1}$ & $89.7$ & $90.5$ & $\textbf{90.1}$ \\ 
			P2, P3, P1,NP1, NP2 & $89.9$  & $90.7$ & $89.0$ & $89.8$ & $89.9$ & $90.2$ & $89.5$ & $89.9$\\ 
			\bottomrule
		\end{tabular}
		}
		
	\end{table}
	
	\begin{table}[bp!]
		\centering
		\caption{Comparison of our model’s performance  with previously published performances on Quora dataset }
		\label{tab:quoraResults}
		
		\begin{tabular}{lc}
			\bottomrule
			Model & Accuracy\\ 
			\midrule
			Wang et al. ($2017$) (Saimese-CNN)  \cite{Wang:2017:BMM:3171837.3171865} & $79.6$  \\
			Wang et al. ($2017$) (Multi-Perspective-CNN)   \cite{Wang:2017:BMM:3171837.3171865} & $81.4$  \\
			Wang et al. ($2017$) (Saimese-LSTM)   \cite{Wang:2017:BMM:3171837.3171865} & $82.58$  \\
			Wang et al. ($2017$) (Multi-Perspective-LSTM)  \cite{Wang:2017:BMM:3171837.3171865} & $83.2$  \\
			Wang et al. ($2017$) (L.D.C)  \cite{Wang:2017:BMM:3171837.3171865} & $85.6$  \\
			Wang et al. ($2017$) (BiMPM)  \cite{Wang:2017:BMM:3171837.3171865} & $88.2$  \\
			Tomar et al. ($2017$) (pt-DECATT$\textsubscript{word}$)   \cite{tomar2017neural} & $87.5$  \\
			Tomar et al. ($2017$) (pt-DECATT$\textsubscript{char}$)   \cite{tomar2017neural} & $88.4$  \\ 
			Lan and Xu ($2018$) (SSE) \cite{lan2018neural} & $87.8$ \\
			\midrule
			Our model & $\textbf{90.3}$  \\ 
			\bottomrule
		\end{tabular}
		
	\end{table}

	Table \ref{tab:quoraResultsProposed} shows the predictive performance of our enhanced paraphrase detection model on the Quora dataset. Performances (accuracy, precision, recall, and F1-score) on test sets are given for different data augmentation steps with learned features and learned plus linguistic features. We first discuss results obtained by using learned features only. Using the original data without any augmentation, our model achieves an accuracy of $89.4\%$. Augmenting the data with step P2   and NP1 (paraphrase and non-paraphrase generation by symmetry) increases the accuracy slightly. Note that this augmentation step has also been performed in earlier works \cite{tomar2017neural, agarwal2018deep}. However, noticeable increase in accuracy is observed when the data is augmented with additional paraphrases using step P2, step P3 (generation by transitive closure), and step P1 (generation by reflexivity), jumping the accuracy to $90.2\%$. We obtain the best performance of $90.3\%$ when in addition to augmenting paraphrases via steps P3, P2, and P1 additional non-paraphrases are generated via step NP1. This is also the current state-of-the-art performance on this dataset. 
	
	Note that by including additional non-paraphrase annotations using step NP2 (generation by negative extension) decreases the accuracy to $89.9\%$ from the high of $90.3\%$ obtained when steps P2, P3, P1, and NP1 are executed. The reason behind this decrease can be determined by analyzing paraphrase concepts (clique in paraphrase graph).  Recall from Section \ref{sc:gen_non_paraphrases} that when even a single edge with label $0$ (a non-paraphrase annotation) between two cliques will generate a complete set of edges between the nodes of the two cliques with label $0$. Thus, any error in such a non-paraphrase annotation gets magnified during the NP2 non-paraphrase augmentation step and degrades the quality of the dataset for paraphrase detection. For example, the incorrect annotation of questions \textit{Is there a way to hack Facebook account ?} and \textit{How can I hack Facebook ?} as non-paraphrase generates numerous erroneous non-paraphrase annotations between paraphrases of the first and second question. Therefore, step NP2 has the potential to degrade paraphrase detection performance when errors exist in non-paraphrase annotations that link large paraphrase concepts. 
	
	When we perform experiments by including linguistic features with learned features, slightly lower performances are obtained.  This highlights that when sufficiently large dataset is available, deep learning models can effectively capture the semantics and contexts of short texts for improved paraphrase detection; for such datasets, the extra effort of including linguistic features is not beneficial. 
	
	Table \ref{tab:quoraResults} presents the performance of previously published work on this dataset. Accuracy values are given in this table because previous works report accuracies only. In \cite{lan2018neural}, $7$ different models are re-implemented on several tasks involving sentence pairs. Quora dataset is used to get results for paraphrase detection task. We only include results of best performing model among all $7$. They find that Shortcut-Stacked Sentence Encoder Model (SSE) \cite{nie2017shortcut} performs the best, giving testing accuracy of $87.8\%$. The previous best accuracy is $88.4\%$ \cite{tomar2017neural}. Our multi-cascaded model beats this result without any data augmentation with an accuracy of $89.4\%$. As seen from the table, our enhanced model outperforms all previous results.  Ensemble {BiMPM} model achieves an accuracy of $88.2\%$ accuracy while pt-DECATT$\textsubscript{char}$ shows a slightly better performance with an accuracy of $88.4\%$. In comparison, our model achieves an accuracy of $90.3\%$, which is almost $2\%$ improvement over the previous best performance. In contrast to \cite{Wang:2017:BMM:3171837.3171865}, we avoid using ensemble model approach (which is computationally costly). Similarly, contrary to results in~\cite{tomar2017neural}, our results are based on word features only and do not use computationally expensive character-based features.

	\subsection{MSRP Dataset}
	
	\begin{table}[!bp]
		\centering
		\caption{Paraphrase detection performance on MSRP dataset }
		\label{tab:msrResultsProposed}
		
		\resizebox{\textwidth}{!}{
		\begin{tabular}{lcccccccccc}
			\toprule
			\multirow{2}{*}{Augmentation} & \multicolumn{4}{c}{Learned Features}  & \multicolumn{4}{c}{Learned + Linguistic Features}\\ 
			\cmidrule(lr){2-5} \cmidrule(l){6-9} 
			& Accuracy & Precision & Recall & F1-score &  Accuracy & Precision & Recall & F1-score \\
			\midrule
			
			None & $74.1$ & $75.3$ & $90.9$ & $ 82.4$ & $ 74.7 $ & $76.2$ &$90.2$& $82.6$ \\
			P2, NP1& $74.8$ & $76.9$ & $88.9$ & $ 82.5$ & $ 75.2 $ &$77.1$&$89.2$& $82.7$ \\
			P2, P3 & $74.4$  & $76.0$ & $89.9$ & $82.3$ & $ 75.0 $ &$76.8$&$89.5$& $82.7$\\
			P2, P3, P1 & $76.8$  &$77.1$& $92.7$ & $84.2$  & $ 77.4 $ &$77.0$&$91.7$& $84.2$\\
			P2, P3, NP1 & $74.4$  &$75.6$&$90.7$& $82.5$  & $ 74.1 $ &$74.7$& $\textbf{92.5}$ & $82.6$\\
			P2, P3, P1, NP1  & $\textbf{77.0}$ &$\textbf{77.3}$&$\textbf{92.7}$ & $\textbf{84.3}$  & $ \textbf{78.3} $ &$\textbf{79.3}$& $91.0$& $ \textbf{84.8}$ \\ 
			P2, P3, P1, NP1, NP2  & $\textbf{77.0}$ &$\textbf{77.3}$&$\textbf{92.7}$ & $\textbf{84.3}$  & $ \textbf{78.3} $ &$\textbf{79.3}$& $91.0$& $ \textbf{84.8}$ \\

			\bottomrule
		\end{tabular}
		}
		
	\end{table}

	\begin{table}[!tp]
		\centering
		\caption{Comparison of our model’s performance with previously published performances on  MSRP dataset }
		\label{tab:msrResults}
		
		\begin{tabular}{lcc}
			\toprule
			Model & Accuracy & F1-score\\ 
			\midrule
			Socher et al. ($2011$) \cite{socher2011dynamic}  & $76.8$ &$83.6$\\
			Madnani et al. ($2012$) \cite{madnani2012re}  & $77.4$ &$84.1$\\
			Ji and Eisenstein (inductive) ($2013$) \cite{ji2013discriminative}  & $77.8$ &$84.3$\\
			Hu et al. ARC-I ($2014$) \cite{hu2014convolutional}  & $69.6$ &$80.3$\\
			Hu et al. ARC-II ($2014$) \cite{hu2014convolutional}  & $69.9$ &$80.9$\\

			El-Alfy et al. ($2015$) \cite{el2015boosting}  & $73.9$ &$81.2$\\
			Kenter and de Rijke ($2015$) \cite{kenter2015short} & $76.6$ &$83.9$\\
			Eyecioglu and Keller ($2015$) \cite{eyecioglu2015twitter}  & $74.4$ &$82.2$\\
			He et al. ($2015$) \cite{he2015multi}  & \textbf{$78.6$} &$84.7$\\

			Dey et al. ($2016$) \cite{dey2016paraphrase}  & $-$ &$82.5$\\
			Wang, Mi et al. ($2016$) \cite{wang2016sentence} & $78.4$ &$84.7$\\
			Yin et al. ($2016$) \cite{yin2016abcnn}  & ${78.9}$ &$\textbf{84.8}$\\
			Pagliardini et al. ($2018$) \cite{N18-1049}  & $76.4$ &$83.4$\\
			Ferreira et al. ($2018$) \cite{ferreira2018combining}  & $74.08$ &$83.1$\\
			Agarwal et al. ($2018$) \cite{agarwal2018deep}  & $77.7$ &$84.5$\\
			Arora and Kansal ($2019$) \cite{arora2019character}  & $\textbf{79.0}$ &$-$\\
			
			\midrule

			Our model & $78.3$ & $\textbf{84.8}$ \\

			\bottomrule
		\end{tabular}
		
	\end{table}
	
	We use the pre-defined split provided in MSRP dataset for training and testing our model. No dev set is provided with this dataset; hence, we hold-out $10\%$ of the training split randomly as dev set. By using a grid search, we find optimal hyper-parameters on this dev set. ELMo embeddings are found to be better for this dataset, while Adam optimizer is selected as optimal one with learning rate of $0.002$. When optimizing dropout rate on this dataset, $0.5$ is found to yield the best results. 
	
	Table \ref{tab:msrResultsProposed} shows the predictive performance of our enhanced paraphrase detection model on MSRP dataset. Performances are given for configurations with and without linguistic features after applying various data augmentation steps. It is observed that without data augmentation, we achieve an F1-score of $82.4\%$. Doubling the data (step P2 and NP1) increases performance slightly in terms of F1-score but the significant gain is obtained when augmenting the data using symmetry, transitivity, and reflexivity (steps P2, P3, P1). This configuration of augmentation gives F1-score of $84.2\%$. The highest F1-score in experiments without any linguistic features of $84.3\%$ is achieved when P2, P3, P1, and NP1 augmentation steps are performed. Please note that adding a pair generated by NP2 augmentation did not affect the performance of the model as it only produces $1$ additional pair (recall Section \ref{MSRP_augmentation}).
	
	Using linguistic features along with learned features boosts the performance of the model. In comparison with experiments without linguistic features, this gain is consistent with the exception of P2, P3 and P1 augmentation scheme, where the F1-score for experiments without and with linguistic features remains the same. Without any augmentation, the F1-score is increased from $82.4\%$ to $82.6\%$ by introducing linguistic features, while with P2 and NP1 augmentation scheme, it is improved from $82.5$ to $82.7$. We achieve maximum performance using data augmentation schemes of P2, P3, P1, and NP1 while using linguistic features. This configuration yields an accuracy of $78.3\%$ and an F1-Score of $84.8\%$, which is the highest among all of our experiments. Note that this augmentation scheme also has the highest F1-score when no linguistic features were used. Adding one additional pair generated by NP2 augmentation did not affect the model's performance.
	
	These results prove that the usefulness of linguistic features is dataset domain and size-specific, and is not generalizable. In particular, the impact of linguistic features is limited when the dataset is large (e.g., Quora dataset) while it is more significant for small datasets (e.g., MSRP dataset).
	
	We compare our best performing variation of experiments with existing state of the art approaches on MSRP dataset in Table \ref{tab:msrResults}. The current state of the art on MSRP is reported to be in \cite{ji2013discriminative}. They report their best results as $80.4\%$ accuracy and an F1-score of $85.9\%$. However, they assume that they have access to testing data at the time of training a model and call it a form of transductive learning. On the other hand, our model is based on inductive learning where training is done in total isolation from the test split. Therefore, it is only fair to compare both models in inductive setup. The training of the model by \cite{ji2013discriminative} in inductive setup yields $77.8\%$ accuracy and F1-score of $84.3\%$ \cite{agarwal2018deep}. These results are far less than what were originally reported in~\cite{ji2013discriminative} using transductive learning. Thus, the current state-of-the-art results in inductive setup are reported by~\cite{he2015multi} and~\cite{wang2016sentence}, with an F1-score of $84.7\%$. Our best model yields an F1-score of $84.8\%$, which is slightly higher than previous state-of-the-art.
	
	In terms of accuracy, the highest performance is reported in~\cite{arora2019character} which is $79.0\%$. However, the F1-score was not reported by the authors. As the class label distribution is highly skewed in this dataset (recall Table~\ref{tab:datasetStats}), the accuracy is not a good measure of performance. In such cases, it is plausible to use F1-score to measure the predictive performance of the models~\cite{sokolova2009systematic}. Therefore, despite higher accuracy reported by the authors, it cannot be concluded that their model yields higher performance than other reported results.

	\subsection{SemEval Dataset}
	In SemEval dataset, the provided dev split is used to fine-tune hyperparameters using grid search. We find that ELMo embeddings yield the best results when Adam is used as optimizer with learning rate $0.002$. The dropout rate is found to be $0.2$. 
	
	\begin{table}[!tp]
		\centering
		\caption{Paraphrase detection performance on SemEval dataset }
		\label{tab:twitterResultsProposed}
		
		\resizebox{\textwidth}{!}{
		\begin{tabular}{lcccccccccc}
			\toprule
			\multirow{2}{*}{Augmentation} & \multicolumn{4}{c}{Learned Features}  & \multicolumn{4}{c}{Learned + Linguistic Features}\\ 
			\cmidrule(lr){2-5} \cmidrule(l){6-9} 
			& Accuracy & Precision & Recall & F1-score &  Accuracy & Precision & Recall & F1-score \\
			\midrule
			
			None & $51.1$ & $27.1$ & $78.9$ & $40.2$ &\textbf{89.0}& $70.8$ & \textbf{80.6} & \textbf{75.4} \\
			P2, NP1 & $78.8$ & $48.5$ & $21.1$ & $29.4$&$88.9$& $\textbf{82.5}$ &$59.4$ & $69.1$  \\
			P2, P3 & $82.5$ & $57.7$ & $60.0$ & $58.9$ & $84.6$ & $63.7$ & $61.1$ & $62.4$ \\
			P2, P3, P1 & $82.2$ & $56.1$ & $68.6$ & $61.7$ & $87.6$ & $77.5$ & $57.1$& $65.8$ \\
			P2, P3, NP1 & $\textbf{84.4}$ & $\textbf{62.6}$ & $62.3$ & $62.5$ & $84.0$ & $60.5$ & $67.4$ & $63.8$ \\
			P2, P3, P1, NP1 &$84.1$ &$60.7$ &$68.0$ &$\textbf{64.2}$&$84.4$& $60.0$ & $76.0$ & $67.0$ \\ 
			P2, P3, P1, NP1, NP2 & $82.7$ & $57.1$ & $\textbf{68.6}$ & $62.3$ &$84.7$& $62.4$ & $67.4$ & $64.8$ \\ 
			
			\bottomrule
		\end{tabular}
		}
		
	\end{table}  
	
	Table \ref{tab:twitterResultsProposed} presents the predictive performance of our model on this dataset. In this table, data augmentation is done with full transitive closure, i.e., $K=*$ for step P3. Without data augmentation and linguistic features, our model achieves an F1-score of $40.2\%$. However, when we apply data augmentation, the F1-score tends to increase. The maximum F1-score without linguistic features is $64.2\%$ which is achieved with P2, P3, P1, and NP1 augmentation. This is consistent with the results of other two datasets used in the study, which also yield maximum performance on this particular augmentation combination. In noisy text, linguistic features affect performance by a large margin. 
	
	Our enhanced model outperforms existing approaches on SemEval dataset in terms of F1-score when no augmentation is done but linguistic features are provided  along with learned features. This particular experiment yields Precision and Recall of $70.8\%$ and $80.6\%$, respectively.  However, it is noticed that precision is lower than recall. When we augment the data using P2 and P3, the precision and recall values tend to become closer, highlighting that model is more robust in distinguishing the two classes. But, the F1-score drops significantly.

		\begin{figure}[!bp]
		\begin{center}

			\includegraphics[page=1, scale=0.52]{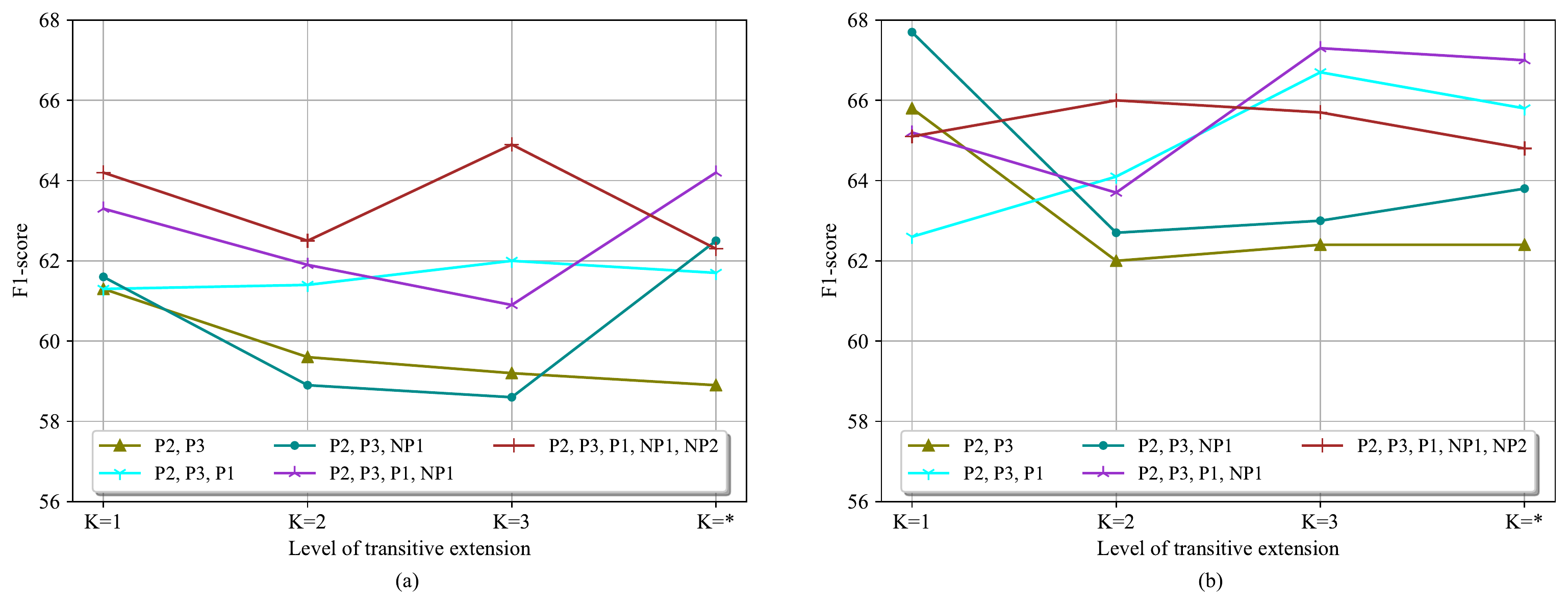}
			\caption{Performance of our model with transitive extensions of order $K=1$,$K=2$,$K=3$, and $K=*$ on SemEval dataset; (a) learned features, (b) learned + linguistic features}
			\label{fig:k_level_compare}
		\end{center}
	\end{figure}

	The reason behind this degradation in performance can be found by investigating pairs generated by transitive closure. It is observed that as we move along the path in graph to find transitive pairs, the meaning of text tends to change, and it becomes more probable that the two documents are no longer paraphrases. This phenomenon is more likely in a noisy short text such as the SemEval dataset. Therefore, instead of applying transitive closure with $K = *$, transitive extension to an order $K$ is plausible. To investigate the effect of $K$ in transitive extension, we perform same experiments with $K = 1$, $K=2$, and $K=3$.  The results of these experiments in terms of F1-score is given in Figure \ref{fig:k_level_compare} (plots (a) and (b) are for without and with linguistic features, respectively). Complete tables of results are included in  \ref{appendix:transitiveExtension}.

	These results show that using linguistic features improves predictive performance as compared to when only learned features are used, and this improvement is consistent for all $K$ and all data augmentation configurations. For order $K$ of transitive extension, we note that moving beyond $K=1$ yield minor improvements in performance. Without any linguistic features, $K=1$ produces the highest F1-score overall, with exception of just one configuration, i.e., P2, P3, P1, NP1, NP2. Similarly, maximum F1-score with linguistic features is also achieved with $K=1$. These observations prove that in noisy text, full transitive closure can produce lower performances and the order $K$ of transitive extension needs to be investigated to determine the optimal data augmentation strategy. Without linguistic features, as opposed to $K=*$, augmenting data with NP2 using $K=1, 2 \text{ or } 3$ does not drop the predictive performance of the model drastically but rather shows a noticeable improvement. However, with inclusion of linguistic features, NP2 augmentation has variable effect on the performance for each order of $K$.

	\begin{table}[tp!]
		\centering
		\caption{Comparison of our model’s performance with previously published performances on SemEval dataset }
		\label{tab:twitterResults}
		
		\begin{tabular}{lccc}
			\toprule
			Model & Precision & Recall & F1-score\\ 
			\midrule
			
			Das and Smith ($2009$) \cite{das2009paraphrase}  & $62.9$ &$63.2$ & $63.0$\\
			Guo and Diab ($2012$) \cite{guo2012modeling}  & $58.3$ &$52.5$ & $65.5$\\
			Ji and Eisenstein ($2013$) \cite{ji2013discriminative}  & $66.4$ &$62.8$ & $64.5$\\
			Xu et al. ($2014$) \cite{xu2014extracting}  & $72.2$ &$72.6$ & $72.4$\\
			Eyecioglu and Keller ($2015$) \cite{eyecioglu2015twitter}  & $68.0$ &$66.9$ & $67.4$\\
			
			Zarella et al. ($2015$) \cite{zarrella2015mitre}  & $56.9$ &$80.6$ & $66.7$\\
			Zhao and Lan ($2015$) \cite{zhao2015ecnu} & $\textbf{76.7}$ &$58.3$ & $66.2$\\
			Vo et al. ($2015$) \cite{vo2015paraphrase}  & $68.5$ &$63.4$ & $65.9$\\
			Karan et al. ($2015$) \cite{karan2015tklbliir}  & $64.5$ &$67.4$ & $65.9$\\

			Dey et al. ($2016$) \cite{dey2016paraphrase}  & $75.6$ &$72.6$ & $74.1$\\
			Huang et al. ($2017$) \cite{huang2017multi}  & $64.3$ &$65.7$ & $65.0$\\
			Lan and Xu ($2018$) \cite{lan2018neural} & $-$ & $-$ & $65.6$ \\
			Agarwal et al. ($2018$) \cite{agarwal2018deep}  & $76.0$ &$74.2$ & $75.1$\\
			
			\midrule
			
			Our model & $70.8$ & $\textbf{80.6}$ & $\textbf{75.4}$ \\

			\bottomrule
		\end{tabular}
		
	\end{table}
	
	We compare our results with existing work in terms of precision, recall and F1-score on test split in Table \ref{tab:twitterResults}. On this dataset, \cite{lan2018neural} in their comparative study of re-implementing $7$ different models, found that Pairwise Word Interaction Model (PWIM) \cite{he2016pairwise} performs the best and achieves F1-score of $65.6\%$, which we report in the table. State of the art performance on the SemEval dataset is reported by \cite{agarwal2018deep} with F1-score of $75.1\%$. Our best performing model outperforms state of the art by achieving an F1-score of $75.4\%$.
	
	\subsection{Impact of Multiple Cascades } \label{subsec:cascadevsdiscriminator}
\begin{figure}[tp!]
		\begin{center}
			\includegraphics[page=1, scale=0.4]{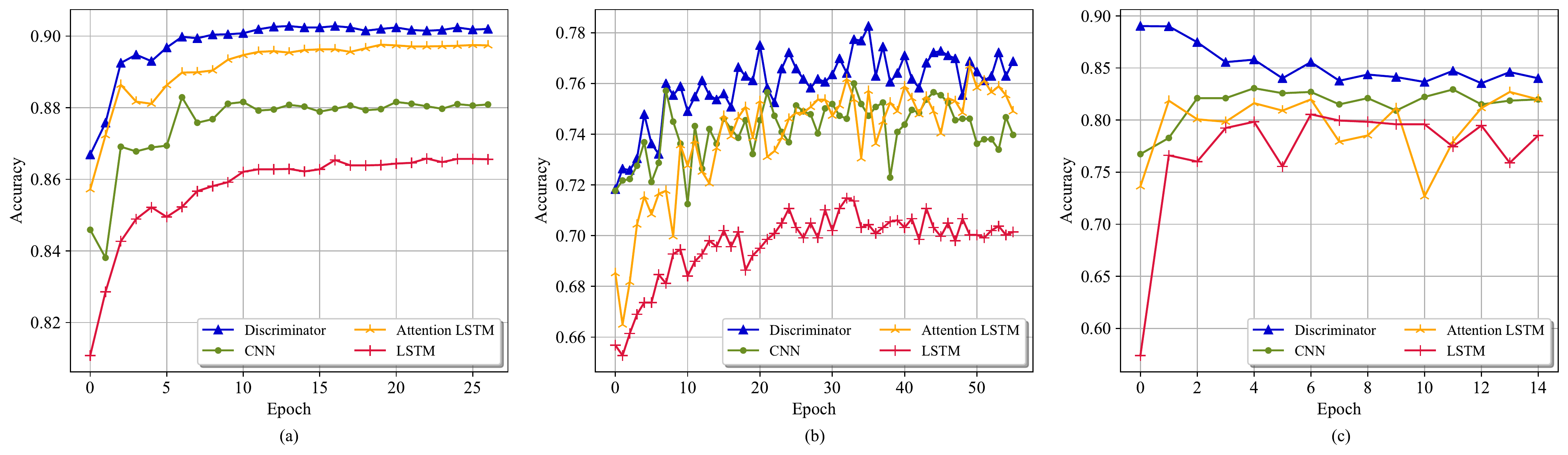}
			\caption{Performance (F1-score or Accuracy) versus training epochs for each feature learning cascades and discriminator network on (a) Quora, (b) MSRP, and (c) SemEval datasets}
			\label{fig:cascadeVsDiscriminator}
		\end{center}
	\end{figure}

	As all cascades are supervised, we can record predictive performances (on test set) for every cascade after each training epoch and compare them with that produced by discriminator network (complete multi-cascaded model). Figure \ref{fig:cascadeVsDiscriminator} shows the performance of every cascade and the complete multi-cascaded model on Quora (plot (a)), MSRP (plot (b)), and SemEval (plot (c)) datasets.
	
	It can be observed that the LSTM-based feature learner yields least performance and remains at the bottom throughout the training, while attention-based convolution feature learner is the next best performer. The attention-based LSTM learner is closely following the discriminator’s performance but discriminator remains at the top during the training process after every epoch. This confirms that training a discriminator based on features learned from multiple perspectives is more fruitful as compared to relying on features learned by only one type of deep learning model. The same trend is followed in all the datasets. These results are presented only for the augmentation configuration which yields best results on each dataset.

	\subsection{Summary} \label{subsec:discussion}
	Our extensive experiments evaluated our enhanced paraphrase detection model along several dimensions. These dimensions include noisy versus clean datasets, large versus small datasets, data augmentation steps and their variations, learned and linguistic features, and cascades in the multi-cascaded model. The key findings of our experiments are summarized below.  
	\begin{enumerate}
		\item Data augmentation improves paraphrase detection predictive performance on all datasets (noisy, clean, large, and small). These easy steps can generate additional annotations that translate into the higher predictive performance from deep learning models.  
		\item Each step for generating additional paraphrase annotations produces an improvement in the prediction performance. On the other hand, only step NP1 for generating additional non-paraphrase annotations consistently improve performance; step NP2 can sometimes cause a decrease in predictive performance especially when the annotation is error-prone (e.g., the notion of paraphrase is not well defined, noisy text). 
		\item Linguistic features are important if a dataset is relatively small and noisy in nature. For such datasets, including linguistic features can produce significant boost in the predictive performance of our model. 
		\item When a dataset is sufficiently large, using linguistic features does not have any effect on the predictive performance. 
		\item For clean text, augmentation scheme of P2, P3, P1, and NP1 gives maximum performance in terms of F1-score on both datasets while for user-generated noisy text, this scheme yields maximum performance only with learned features only. When linguistic features are used, maximum performance is achieved without any augmentation.
		\item It is not recommended to use full transitive closure ($K=*$) for user-generated noisy datasets as no noticeable and consistent improvement in performance is observed. For such datasets, data augmentation with the transitive extension should be investigated. 
	\end{enumerate}

	\section{Conclusion } \label{sec:Conclusion}
	We present a data augmentation strategy and a multi-cascaded model for enhanced paraphrase detection. The data augmentation strategy generates additional paraphrase and non-paraphrase annotations based on the graph analysis of the existing annotations. This strategy is easy to implement and yields significant improvement in the performance of deep learning models for paraphrase detection. The multi-cascaded model employs multiple feature learners to encode and classify short text pairs. As such, it exploits multiple semantic cues to distinguish between paraphrases and non-paraphrases. The proposed multi-cascaded model is both deep and wide in architecture, and it embodies previous best practices in deep models for paraphrase detection. 
	
	We evaluate our enhanced model on three benchmark datasets representing noisy and clean test. Our model produces a higher predictive performance on all three datasets beating all previously published results on them. We also study the impact of different steps in data augmentation, the use of linguistic features in conjunction with learned features, and different deep models. The results show that data augmentation is generally beneficial and linguistic features only help for small and noisy text datasets. Furthermore, it is seen that multiple models can boost predictive performance beyond that achievable from any single model.  
	
	This work provides a comprehensive treatment of paraphrase detection that includes small and large datasets, clean and noisy texts, CNN and LSTM-based models, learned features, hand-crafted linguistic features, and a new data augmentation strategy. In the future, it would be beneficial to investigate strategies for resolving conflicts and achieving quality annotations in noisy data.

	\appendix
	\section{Effect of Transitive Extension Levels on SemEval Dataset} \label{appendix:transitiveExtension}
	\begin{table}[h!]
		\centering
		\caption{Paraphrase detection performance on SemEval dataset with K=1 }
		\label{tab:twitterResultsk1}
		
		\resizebox{\textwidth}{!}{
		\begin{tabular}{lcccccccccc}
			\toprule
			\multirow{2}{*}{Augmentation} & \multicolumn{4}{c}{Learned Features}  & \multicolumn{4}{c}{Learned + Linguistic Features}\\ 
			\cmidrule(lr){2-5} \cmidrule(l){6-9} 
			& Accuracy & Precision & Recall & F1-score &  Accuracy & Precision & Recall & F1-score \\
			\midrule
			
			P2, P3 & $82.8$ & $57.9$ & $65.1$ & $61.3$ & $86.8$ & $71.3$ & $61.1$ & $65.8$ \\
			P2, P3, P1 & $82.1$ & $55.9$ & $68.0$ & $61.3$ & $85.3$ & $66.9$ & $58.9$ & $62.6$\\
			P2, P3, NP1 & $81.9$ & $55.2$ & $69.7$ & $61.6$ & $\textbf{87.4}$ & $\textbf{72.5}$ & $63.4$ & $\textbf{67.7}$\\
			P2, P3, P1, NP1 & $82.8$ & $57.1$ & $\textbf{70.9}$ & $63.3$ & $86.8$ & $72.2$ & $59.4$ & $65.2$ \\ 
			P2, P3, P1, NP1, NP2 & $\textbf{85.2}$ & $\textbf{64.9}$ & $63.4$ & $\textbf{64.2}$ & $84.6$ & $61.9$ & $\textbf{68.6}$ & $65.1$\\

			\bottomrule
		\end{tabular}
		}
		
	\end{table}
	
	\begin{table}[h!]
		\centering
		\caption{Paraphrase detection performance on SemEval dataset with K=2 }
		\label{tab:twitterResultsk2}
		
		\resizebox{\textwidth}{!}{
		\begin{tabular}{lcccccccccc}
			\toprule
			\multirow{2}{*}{Augmentation} & \multicolumn{4}{c}{Learned Features}  & \multicolumn{4}{c}{Learned + Linguistic Features}\\ 
			\cmidrule(lr){2-5} \cmidrule(l){6-9} 
			& Accuracy & Precision & Recall & F1-score &  Accuracy & Precision & Recall & F1-score \\
			\midrule
			
			P2, P3 & $82.2$ & $56.7$ & $62.9$ & $59.6$ & $85.0$ & $65.6$ & $58.9$ & $62.0$\\
			P2, P3, P1 & $81.1$ & $53.6$ & $\textbf{72.0}$ & $61.4$ & $83.3$ & $58.1$ & $71.4$ & $64.1$\\
			P2, P3, NP1 & $82.0$ & $56.3$ & $61.7$ & $58.9$ & $85.1$ & $65.6$ & $60.0$ & $62.7$\\
			P2, P3, P1, NP1 & $83.5$ & $59.9$ & $64.0$ & $61.9$ & $\textbf{85.4}$ & $\textbf{66.5}$ & $61.1$ & $63.7$\\ 
			P2, P3, P1, NP1, NP2 & $\textbf{84.2}$ & $\textbf{62.1}$ & $62.9$ & $\textbf{62.5}$ & $83.2$ & $57.1$ & $\textbf{78.3}$ & $\textbf{66.0}$\\

			\bottomrule
		\end{tabular}
		}
		
	\end{table}
	
	\begin{table}[h!]
		\centering
		\caption{Paraphrase detection performance on SemEval dataset with K=3 }
		\label{tab:twitterResultsk3}
		\resizebox{\textwidth}{!}{
		\begin{tabular}{lcccccccccc}
			\toprule
			\multirow{2}{*}{Augmentation} & \multicolumn{4}{c}{Learned Features}  & \multicolumn{4}{c}{Learned + Linguistic Features}\\ 
			\cmidrule(lr){2-5} \cmidrule(l){6-9} 
			& Accuracy & Precision & Recall & F1-score &  Accuracy & Precision & Recall & F1-score \\
			\midrule
			
			P2, P3 & $82.6$ & $57.9$ & $60.6$ & $59.2$ & $86.0$ & $63.7$ & $61.1$ & $62.4$\\
			P2, P3, P1 & $81.7$ & $54.8$ & $\textbf{71.4}$ & $62.0$ & $86.3$ & $67.6$ & $65.7$ & $66.7$\\
			P2, P3, NP1 & $85.0$ & $69.0$ & $50.9$ & $58.6$ & $81.5$ & $54.1$ & $\textbf{75.4}$ & $63.0$\\
			P2, P3, P1, NP1 & $\textbf{86.0}$ & $\textbf{73.4}$ & $52.0$ & $60.9$ & $\textbf{86.8}$ & $\textbf{69.5}$ & $65.1$ & $\textbf{67.3}$\\ 
			P3, P2,  P1, NP1, NP2 & $85.6$ & $65.9$ & $64.0$ & $\textbf{64.9}$ & $85.6$ & $65.2$ & $66.3$ & $65.7$ \\

			\bottomrule
		\end{tabular}
		}
		
	\end{table}

	\newpage

\nolinenumbers
\bibliography{mybibfile}
	
\end{document}